\documentclass[letterpaper, 10 pt, conference]{ieeeconf}  
\IEEEoverridecommandlockouts
\title{
    \LARGE \bf Learning Direct Control from Optimization \\ for Dynamic Target Tracking
}

\author{
    Bing Sun, Yue Lin, Canhui Wen, Xin Chen, Yang Liu, Dong Wang*, Huchuan Lu
}

\usepackage{graphicx}
\usepackage{amsmath}
\usepackage{amssymb}
\usepackage{booktabs}
\usepackage{multirow}
\usepackage{tabularx}
\usepackage{makecell}
\usepackage{array}
\usepackage{cite}

\usepackage[linesnumbered, ruled]{algorithm2e}
\makeatletter
\renewcommand{\@algocf@pre@ruled}
\makeatother

\begin{document}

\thispagestyle{empty}
\pagestyle{empty}

\maketitle

\begin{abstract}
Dynamic target tracking requires a mobile robot to maintain target visibility while navigating obstacles with smooth, feasible controls. Online trajectory optimization explicitly encodes these objectives but requires repeated numerical solving, whereas waypoint-based neural policies separate geometric prediction from control execution. We introduce LECO, a framework that transfers tracking optimization into offline learning of a direct-control policy. LECO predicts control sequences and maps them to trajectories through a differentiable kinematic model, allowing trajectory-level objectives to directly guide control learning. A decaying expert-guidance curriculum initializes effective tracking behavior and progressively shifts learning toward these task objectives. At deployment, a predictive safety lock checks short-horizon obstacle clearance and invokes a classical planning and control fallback when the predicted motion is unsafe or the target prediction is invalid. Across three simulated obstacle densities, LECO reduces root-mean-square linear jerk by 14.1--31.2\% relative to the best competing method for this metric and achieves the highest tracking success rate, with per-cycle computation times of 4.25--5.63\,ms. Experiments on two robot platforms further demonstrate tracking through indoor narrow passages and outdoor forest environments. These results demonstrate that learning direct control from optimization enables smooth and robust dynamic target tracking.
\end{abstract}

\section{Introduction}

Dynamic target tracking requires a mobile robot to keep a moving target in view while navigating the surrounding environment~\cite{40,44}. In cluttered scenes, these requirements are tightly coupled: a collision-free maneuver may place an obstacle between the robot and the target, while maintaining visibility may require turning through narrow passages. The robot must therefore coordinate target distance, line of sight (LoS), obstacle clearance, and control smoothness under its own kinematic constraints. The central challenge is to translate these spatial tracking objectives into feasible control actions as both the target and the local environment change.

Trajectory optimization provides an explicit way to express these objectives. Optimization-based trackers account for target motion, visibility, and obstacle avoidance when generating robot trajectories~\cite{5,6,47}, while model predictive control (MPC) incorporates motion constraints into control computation~\cite{4}. This approach requires numerical solving as observations change. Learning-based planners offer a complementary route by transferring trajectory costs into network training~\cite{8,10}. For example, YOPOv2-Tracker learns motion-primitive offsets from trajectory costs and converts the resulting predictions into quadrotor commands~\cite{8}. Kinematics-aware learning also incorporates motion models into planning~\cite{43,48}. These developments motivate learning tracking behavior directly in the control space of a mobile robot, so that visibility, clearance, and smoothness objectives shape the commands through the motion they produce.

Our key idea is to retain the explicit objectives of tracking optimization while moving their optimization into offline learning of a direct-control policy. A differentiable kinematic model provides the connection: it maps a predicted control sequence to a robot trajectory and propagates trajectory-level costs back to the policy. Target visibility and obstacle clearance can then guide control learning together with control smoothness, using the platform's motion model throughout. This formulation connects what the robot should achieve in the environment with the controls it will execute.

\begin{figure}[t]
    \vspace{2 mm}
    \centering
    \includegraphics[width=\linewidth]{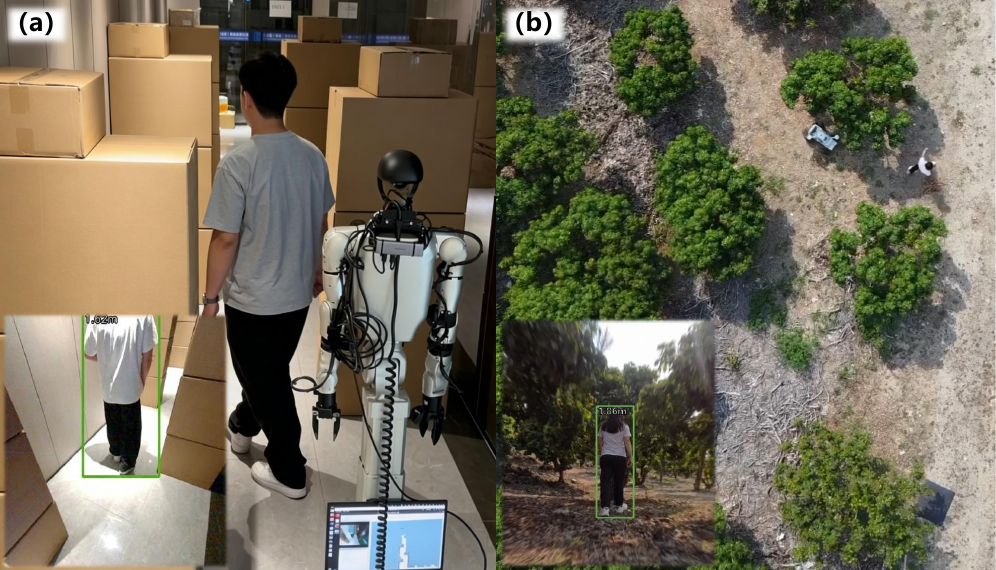}
    \caption{Real-world target-tracking experiments on two robot platforms. (a) Unitree G1D tracking in an indoor narrow-passage environment. (b) AgileX HUNTER 2.0 tracking in an outdoor forest. The bottom-right inset in each panel shows the robot's first-person view with the estimated target distance.
    }
    \vspace{-4 mm}
    \label{fig:head}
\end{figure}

\begin{figure*}[t]
    \vspace{2 mm}
    \centering
    \includegraphics[width=\linewidth]{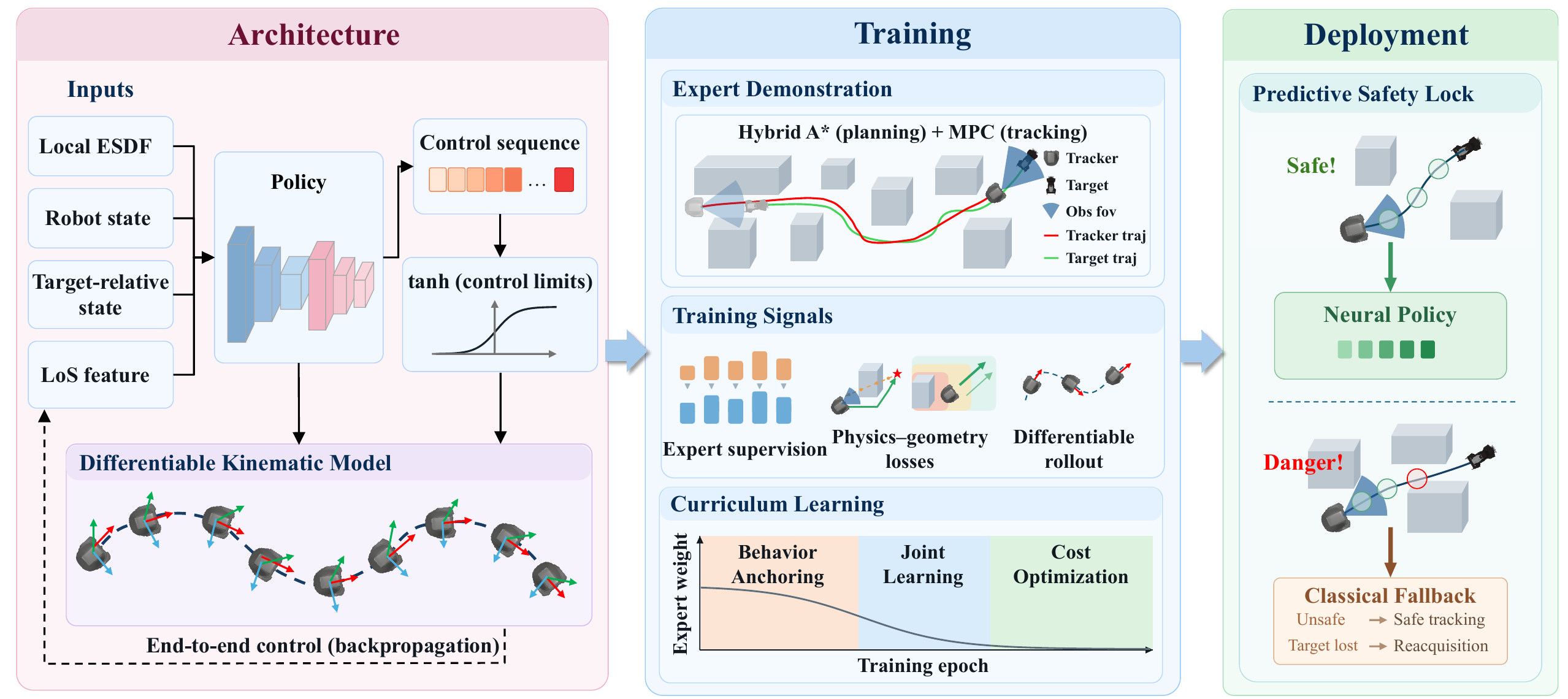}
    \caption{Overview of LECO, a kinematic-integrated neural control framework. It transfers optimization objectives into policy learning and combines learned control with predictive safety protection. (1) Kinematic-integrated control directly predicts future controls and rolls them out through a differentiable kinematic model. (2) Optimization-to-policy learning transfers expert behavior and physics--geometry objectives into the neural policy. (3) A predictive safety lock enables neural execution under safe conditions and switches to the classical fallback branch when necessary.}
    \vspace{-3 mm}
    \label{fig:framework}
\end{figure*}

We instantiate this idea in \textbf{LECO}, a framework for learning direct control from optimization for dynamic target tracking (Fig.~\ref{fig:framework}). Given local geometry, robot motion, and target-relative observations, LECO predicts a future control sequence and rolls it out through a platform-specific differentiable kinematic model. Tracking objectives evaluated along this rollout jointly encourage the desired target distance, clear lines of sight, obstacle clearance, and smooth motion. A decaying expert-guidance curriculum establishes effective tracking behavior early in training, then progressively shifts learning toward these task objectives. The expert thus provides an initialization for control learning, while the final policy is refined by the quality of the motion it predicts.

At deployment, LECO executes the first predicted control in a receding-horizon manner, avoiding iterative task optimization in the neural branch. A predictive safety lock checks obstacle clearance along the short-horizon rollout before execution. When the predicted motion is unsafe or the target prediction is invalid, it invokes a classical hybrid A*~\cite{3} and MPC~\cite{4} branch for obstacle avoidance or target reacquisition. Neural execution resumes once the target prediction and clearance checks permit it and the lock duration has elapsed. This deployment strategy combines learned direct control with explicit motion checks and a fallback that continues to serve the tracking task.

Experiments validate this formulation for smooth and robust tracking. Across three simulated obstacle densities, LECO achieves the lowest root-mean-square linear jerk and the highest tracking success rate among the evaluated methods, with per-cycle computation times of 4.25--5.63\,ms. In the 150-obstacle setting, it achieves an 86\% success rate and a 6.32\% occlusion rate. Further evaluations in dense layouts and temporary obstacle disturbances show that the neural policy supplies 88.57--95.81\% of control decisions. Real-world experiments with platform-specific policies on the differential-drive Unitree G1D and Ackermann-steered AgileX HUNTER 2.0 demonstrate tracking through indoor narrow passages and outdoor forest environments (Fig.~\ref{fig:head}).

Our main contributions are as follows:
\begin{itemize}
    \item We introduce LECO, a direct-control learning framework for dynamic target tracking that propagates trajectory-level tracking objectives through a differentiable kinematic rollout to the control policy.
    \item We develop a decaying expert-guidance curriculum for visibility-aware control learning and a predictive safety lock that combines neural execution with classical fallback for obstacle avoidance and target reacquisition.
    \item We demonstrate improved control smoothness and tracking success across three simulated obstacle densities, and validate the framework on two robot platforms in indoor and outdoor environments.
\end{itemize}

The source code and implementation details will be made publicly available.

\section{Related Work}

\subsection{Classical Optimization-Based Target Tracking}

Traditional planning methods include MPC, MPPI, ESDF-based nonlinear optimization and so on~\cite{23,24,35,36}. They typically encode line-of-sight maintenance and obstacle clearance as cost terms for dynamic obstacle avoidance and target tracking. However, in nonconvex scenarios involving narrow passages and tight corners, online solvers such as SQP and L-BFGS are prone to local minima and zero-velocity deadlocks~\cite{37}. The computational burden depends strongly on the solver, optimization horizon, and environment complexity. Online optimization is still required at every control cycle and becomes more expensive as problem complexity grows. Meanwhile, modeling and state-prediction errors can accumulate over long-horizon rollouts. 

\subsection{Learning-Based Motion Planning and Control}

To avoid repeated online optimization, DRL and behavioral cloning have been widely used in robotic motion planning~\cite{30,31,41}. Recent cost-guided methods further train policies with trajectory-level objectives. For example, YOPOv2-Tracker~\cite{8} back-propagates trajectory costs for target tracking, but still predicts intermediate motion representations before execution. Such waypoint- or trajectory-based architectures separate geometric prediction from platform control~\cite{7,8,9}. This separation can introduce a representation mismatch between training objectives and executable controls~\cite{13,14,16}. LECO follows the cost-guided learning paradigm, but directly predicts control sequences and introduces explicit target-tracking and LoS objectives.

\subsection{Optimization Distillation and Safe Deployment}

Optimization-to-Policy transfers the knowledge encoded in traditional optimizers, such as MPC, into neural policies~\cite{26,43}. It provides a promising way to combine physical constraints with efficient inference. However, existing optimization-distillation methods typically mainly focus on generic motion planning or point-to-point navigation~\cite{7}. Direct cost optimization can also converge to near-zero controls during early training. Moreover, black-box policies lack deterministic safety guarantees in out-of-distribution (OOD) scenarios~\cite{11,17,22,50}. Control barrier function (CBF)-based safety filters add an online constrained safety-filtering step~\cite{51}. In contrast, fallback strategies, such as emergency stoppage or global replanning, can disrupt the continuity of dynamic tracking.

\section{Methodology}

\subsection{Framework Overview}

Traditional planning and control methods (as shown in Fig.~\ref{fig:method}(a)) can be formulated as a platform-specific solver $\mathcal{S}_r$. The solver generates a feasible control sequence online through search or iterative optimization:
\begin{equation}
\mathbf{U}_t^{r,\mathrm{trad}}
=
\mathcal{S}_r
\left(
\mathbf{o}_t^r,
\hat{\mathbf{P}}_{\mathrm{tar},t},
\mathcal{M}_t;
\boldsymbol{\eta}_r
\right),
\label{eq:traditional_solver}
\end{equation}

Here, \(\hat{\mathbf P}_{\mathrm{tar},t}\) denotes the predicted target trajectory, \(\mathcal M_t\) denotes the environment representation used by the traditional solver, $\boldsymbol{\eta}_r$ denotes the platform-specific motion and safety parameters.

LECO replaces this online optimization with offline policy learning. The policy directly predicts a future control sequence. A differentiable kinematic model then rolls out the corresponding state trajectory:

\begin{equation}
\begin{gathered}
\hat{\mathbf{U}}_t^r
=
\pi_{\theta_r}(\mathbf{o}_t^r),
\quad
\hat{\mathbf{X}}_t^r
=
\Phi_r
\left(
\mathbf{x}_t^r,
\hat{\mathbf{U}}_t^r
\right),\\
\mathbf o_t^r
=
\left\{
\mathbf E_t^{\mathrm{loc}},
\mathbf s_t^{\mathrm{rob}},
\mathbf s_{\mathrm{tar},t}^{\mathrm{rel}},
\mathbf r_t^{\mathrm{vis}}
\right\}.
\end{gathered}
\end{equation}

where \(\mathbf E_t^{\mathrm{loc}}\), \(\mathbf s_t^{\mathrm{rob}}\), \(\mathbf s_{\mathrm{tar},t}^{\mathrm{rel}}\), and \(\mathbf r_t^{\mathrm{vis}}\) denote the local ESDF, robot motion state, target-relative state, and LoS features. The policy is trained with curriculum-based expert supervision and physics- and geometry-driven differentiable losses.

During deployment, the policy operates in a receding-horizon manner and executes only the first control action. A predictive safety lock checks the ESDF clearance over a short prediction horizon. If the target prediction becomes invalid or the predicted trajectory becomes unsafe, the system switches to the classical hybrid A* and MPC branch.

\subsection{Kinematic-Integrated Control Policy}

\begin{figure}[t]
    \centering
    \vspace{2 mm}
    \includegraphics[width=\linewidth]{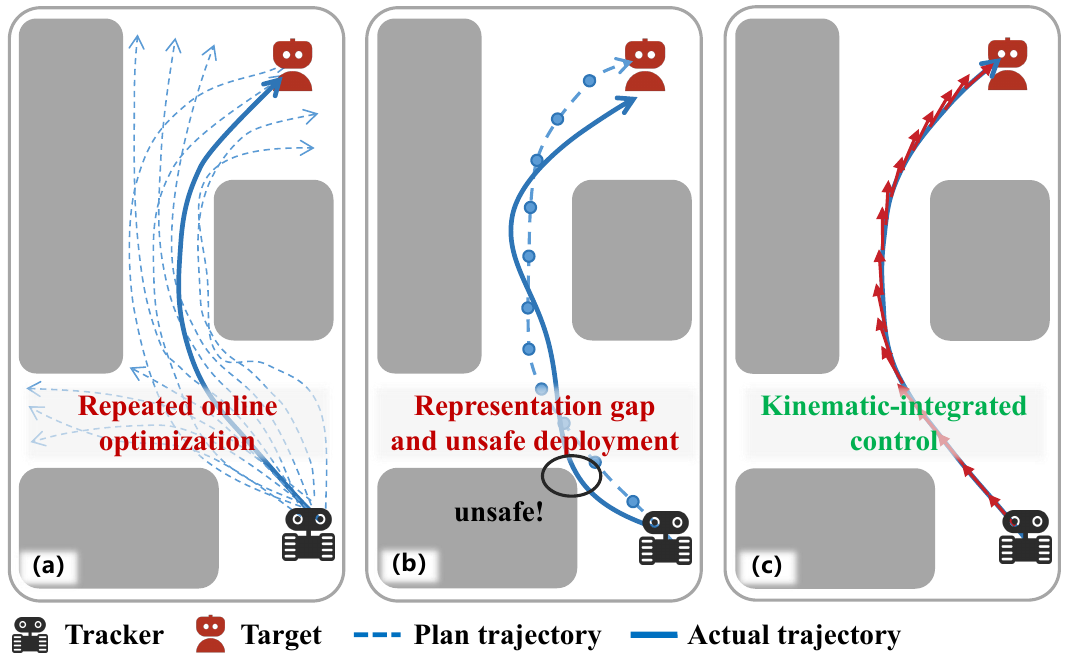}
    \caption{Comparison between existing methods and the proposed LECO. (a) Traditional hybrid A* and MPC-based planning repeatedly solves planning and control problems online. (b) Waypoint-based neural planning suffers from a representation gap between planning and control. (c) LECO employs kinematic-integrated direct control for end-to-end target tracking.}
    \label{fig:method}
    \vspace{-3 mm}
\end{figure}

For robot platform \(r\), LECO uses a platform-specific control head and differentiable kinematic model. We adopt the differential-drive and Ackermann kinematic models from~\cite{45} and~\cite{46}, respectively. 
To model the dynamic target motion, we estimate a degree-\(2m\) B\'ezier trajectory from the latest \(m+1\) observed positions.
Let $T=m\tau$, where $\tau$ is the observation interval.
The time variable $\xi\in[-T,T]$ is normalized as
$s(\xi)=(\xi+T)/(2T)$.
The target trajectory and its current motion state are then estimated as
\begin{equation}
\hat{\mathbf p}_{\mathrm{tar}}(\xi)
=
\sum_{j=0}^{2m}
\mathbf c_j
B_j^{2m}\!\left(s(\xi)\right),
\end{equation}

\begin{equation}
\mathbf p_{\mathrm{tar},t}
=
\hat{\mathbf p}_{\mathrm{tar}}(\delta_t),
\quad
\mathbf v_{\mathrm{tar},t}
=
\dot{\hat{\mathbf p}}_{\mathrm{tar}}(\delta_t),
\label{eq:target_prediction}
\end{equation}
where $\mathbf c_j$ denotes the B\'ezier control points.
The historical observations provide position constraints, while the minimum-acceleration criterion is used to determine the remaining degrees of freedom.
Here, \(\xi=0\) corresponds to the latest trajectory update, and \(\delta_t\) denotes the elapsed time from this reference to the current control cycle.

Using the angular encoding \(\mathbf e(\phi)=[\sin\phi,\cos\phi]^{\mathsf T}\),
the predicted target state is then expressed in the robot-relative frame as
\begin{equation}
\mathbf s_{\mathrm{tar},t}^{\mathrm{rel}}=
\left[
\rho_t,\,
\mathbf e^{\mathsf T}(\phi_t),\,
\mathbf e^{\mathsf T}(\phi_{v,t}),\,
\|\mathbf v_{\mathrm{tar},t}\|_2
\right]^{\mathsf T},
\label{eq:relative_target_state}
\end{equation}
where \(\rho_t\), \(\phi_t\), and \(\phi_{v,t}\) denote the target distance, relative bearing, and target velocity direction, respectively, in the robot frame.

Unlike methods that first predict geometric waypoints and then convert them into control commands (as shown in Fig.~\ref{fig:method}(b)), LECO directly predicts a future control sequence from the current observation (as shown in Fig.~\ref{fig:method}(c)):
\begin{equation}
\hat{\mathbf U}_t^r
=
\pi_{\theta_r}(\mathbf o_t^r)
=
\left[
\hat{\mathbf u}_{t|t}^r,
\ldots,
\hat{\mathbf u}_{t+H-1|t}^r
\right]^{\mathsf T}.
\label{eq:control_policy}
\end{equation}
The policy takes an \(80\times80\) robot-centered local ESDF (\(\sim4\,\mathrm{m}\times4\,\mathrm{m}\)), current linear and angular velocities, a 6-D target-relative state, and eight LoS ESDF samples, and predicts a 30-step control sequence. We use $H=30$ and $\Delta t=0.1\,\mathrm{s}$, corresponding to a
$3.0\,\mathrm{s}$ prediction horizon. The policy runs at $10\,\mathrm{Hz}$
in a receding-horizon manner and executes only the first predicted control.

The policy outputs are first mapped to the platform-specific control limits using differentiable \(\tanh\) scaling. For the differential-drive platform, the controls are rolled out as
\begin{equation}
\begin{aligned}
\hat x_{t+k+1|t}^r
&=
\hat x_{t+k|t}^r
+
\Delta t\,\hat v_{t+k|t}^r
\cos\hat\theta_{t+k|t}^r,\\
\hat y_{t+k+1|t}^r
&=
\hat y_{t+k|t}^r
+
\Delta t\,\hat v_{t+k|t}^r
\sin\hat\theta_{t+k|t}^r,\\
\hat\theta_{t+k+1|t}^r
&=
\hat\theta_{t+k|t}^r
+
\Delta t\,\hat\omega_{t+k|t}^r,
\end{aligned}
\label{eq:kinematic_rollout}
\end{equation}
with $\hat{\mathbf x}_{t|t}^r=\mathbf x_t^r$ and k=0,\ldots,H-1. Equation~\eqref{eq:kinematic_rollout} explicitly defines the differentiable rollout operator \(\Phi_r\). For Ackermann platforms, the rollout is replaced by the corresponding differentiable Ackermann model~\cite{46}.

This differentiable rollout allows spatial objectives, including target tracking and LoS constraints, to backpropagate directly to the control policy.

\subsection{Cost-Guided Optimization-to-Policy Learning}

Inspired by the cost-guided learning strategy of YOPOv2-Tracker~\cite{8}, LECO back-propagates differentiable task costs through the kinematic rollout in Eq.~\eqref{eq:kinematic_rollout}. Unlike YOPOv2, LECO applies these costs directly to predicted control sequences. We constrain only planar motion and control inputs; low-level execution is handled by the platform controller.

\subsubsection{Traditional Costs as Differentiable Losses}

Following YOPOv2-Tracker~\cite{8}, the base loss $\mathcal L_{\mathrm{base}}^r$ includes collision, control-feasibility, and smoothness terms. We further introduce target-visibility and low-clearance escape losses: 
\begin{equation}
\mathcal L_{\mathrm{cost}}^r
=
\mathcal L_{\mathrm{base}}^r
+
\mathcal L_{\mathrm{tv}}^r
+
\lambda_{\mathrm{out}}
\mathcal L_{\mathrm{out}}^r,
\label{eq:cost_loss}
\end{equation}
where \(\mathcal L_{\mathrm{tv}}^r\) denotes the target--visibility loss, and \(\mathcal L_{\mathrm{out}}^r\) denotes the low-clearance ESDF gradient-escape loss. We further introduce soft line-of-sight gating and ESDF gradient guidance for low-clearance regions.

\noindent\textit{a) Soft LoS Gating and Dynamic Extrapolation:}
For prediction step \(k\), the target position is taken from the predicted trajectory when available; otherwise, constant-velocity extrapolation is used. Define
\begin{equation}
\mathbf p_{\mathrm{tar},k}^{\mathrm{cv}} =
\mathbf p_{\mathrm{tar},0} +
k\Delta t\,\mathbf v_{\mathrm{tar},0}.
\end{equation}
The target position is then given by
\begin{equation}
\hat{\mathbf p}_{\mathrm{tar},k}
=\mu_k\mathbf p_{\mathrm{tar},k}^{\mathrm{pred}}
+
(1-\mu_k)\mathbf p_{\mathrm{tar},k}^{\mathrm{cv}},
\label{eq:target_fallback}
\end{equation}
where \(\mu_k\in\{0,1\}\) indicates whether a valid target prediction is available. For brevity, let
\(\mathbf p_k^r=[\hat x_{t+k|t}^r,\hat y_{t+k|t}^r]^{\mathsf T}\)
denote the predicted robot position at rollout step \(k\). We sample \(K\) points along the robot-target line segment:
\begin{equation}
\mathbf z_{k,j}
=
(1-\alpha_j)\mathbf p_k^r
+
\alpha_j\hat{\mathbf p}_{\mathrm{tar},k},
\label{eq:los_samples}
\end{equation}
where $\alpha_j=j/(K+1)$, $j=1,\ldots,K$, and $K=8$.

The minimum LoS clearance is used to construct a soft visibility gate:
\begin{equation}
\begin{gathered}
c_k^{\mathrm{los}}
=
\min_{1\leq j\leq K}
\mathcal D_{\mathrm{ESDF}}(\mathbf z_{k,j}),
\\
g_k^{\mathrm{vis}}
=
\sigma
\left(
\frac{
\operatorname{sg}(c_k^{\mathrm{los}})
-
d_{\mathrm{vis}}^r
}{
T_{\mathrm{vis}}
}
\right),
\label{eq:visibility_gate}
\end{gathered}
\end{equation}
where $\operatorname{sg}(\cdot)$ denotes the standard stop-gradient operator. It prevents the visibility gate from directly changing the LoS clearance gradient. The gate therefore only modulates the target-distance objective.

Let
$d_k^{\mathrm{tar}}
=
\|\mathbf p_k^r-\hat{\mathbf p}_{\mathrm{tar},k}\|_2$
and
$w_k=(H-k+1)/H$.
The target-tracking and LoS objectives are combined as

\begin{equation}
\mathcal L_{\mathrm{tv}}^r
=
\frac{1}{H}\sum_{k=1}^{H}
\left(
\lambda_{\mathrm{dist}}\ell_k^{\mathrm{dist}}
+
\lambda_{\mathrm{min}}\ell_k^{\mathrm{min}}
+
\lambda_{\mathrm{los}}\ell_k^{\mathrm{los}}
\right),
\label{eq:target_visibility_loss}
\end{equation}

and where
\begin{equation}
\begin{gathered}
\ell_k^{\mathrm{dist}}
=
w_k g_k^{\mathrm{vis}}
\frac{(d_k^{\mathrm{tar}}-d_{\mathrm{des}})^2}
{\max(d_k^{\mathrm{tar}},\epsilon)},\\
\ell_k^{\mathrm{min}}
=
[d_{\mathrm{min}}^r-d_k^{\mathrm{tar}}]_+^2,\\
\ell_k^{\mathrm{los}}
=
\left(
\sum_{j=1}^{K}
[d_{\mathrm{los}}
-\mathcal D_{\mathrm{ESDF}}(\mathbf z_{k,j})]_+
\right)^2.
\end{gathered}
\end{equation}

Here, $[a]_+=\max(a,0)$. Bilinear interpolation is used for ESDF queries, allowing these spatial constraints to backpropagate directly through the predicted trajectory.

\noindent\textit{b) Low-Clearance ESDF Gradient Escape:}
Low-clearance regions may induce near-zero-velocity local optima. 
We therefore define a clearance-aware weight as
\begin{equation}
g_k^d
=
\operatorname{sg}\left[
\min\left(
\left[
1-
\frac{\mathcal D_{\mathrm{ESDF}}(\mathbf p_k^r)}
{d_{\mathrm{danger}}^r}
\right]_+,
g_{\max}
\right)
\right].
\label{eq:danger_weight}
\end{equation}

The predicted motion direction and the normalized ESDF ascent direction are defined as

\begin{equation}
\begin{gathered}
\hat{\mathbf m}_k
=
\frac{
\mathbf p_{k+1}^r-\mathbf p_k^r
}{
\|\mathbf p_{k+1}^r-\mathbf p_k^r\|_2+\epsilon
},
\\[2pt]
\hat{\mathbf g}_k
=
\operatorname{sg}\left(
\frac{
\nabla_{\mathbf p}\mathcal D_{\mathrm{ESDF}}(\mathbf p_k^r)
}{
\|\nabla_{\mathbf p}\mathcal D_{\mathrm{ESDF}}(\mathbf p_k^r)\|_2+\epsilon
}
\right).
\end{gathered}
\label{eq:escape_directions}
\end{equation}

We then encourage the predicted motion to align with the direction of increasing ESDF clearance:
\begin{equation}
\mathcal L_{\mathrm{out}}^r
=
\frac{1}{H}
\sum_{k=0}^{H-1}
g_k^d
\left[
\delta_{\mathrm{out}}
-
\hat{\mathbf m}_k^{\mathsf T}
\hat{\mathbf g}_k
\right]_+.
\label{eq:gradient_escape_loss}
\end{equation}
Here, $\delta_{\mathrm{out}}$ specifies the required directional alignment. The stop-gradient operator makes the danger weight and ESDF ascent direction local guidance signals, avoiding gradients through the gating weight or higher-order ESDF derivatives.

\subsubsection{Curriculum Expert-to-Cost Knowledge Transfer}

Early in training, collision, clearance, and smoothness objectives may produce near-zero controls. To avoid this collapse, hybrid A* and MPC provide expert controls. A decaying expert anchor then shifts the policy from behavior imitation to physics- and geometry-driven cost optimization.

\begin{figure}[t]
    \centering
    \vspace{2 mm}
    \includegraphics[width=\linewidth]{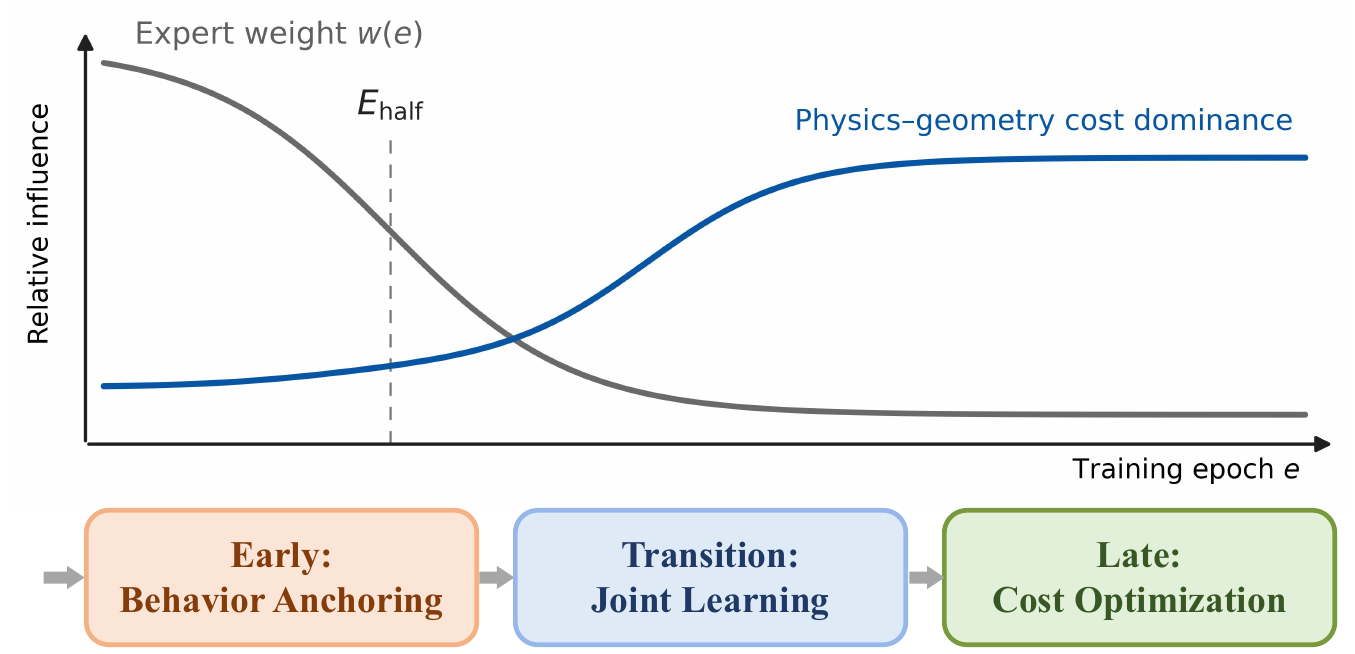}
    \caption{Curriculum expert-to-cost knowledge transfer. The decaying expert weight $w(e)$ guides training from early behavior anchoring to joint learning and finally to physics- and geometry-driven cost optimization, preventing zero-control collapse while gradually reducing expert dependence.}
    \vspace{-2 mm}
    \label{fig:method2}
\end{figure}

To reduce the effect of different control scales across platforms, we normalize the predicted and expert controls using the same platform-specific limits:
\begin{equation}
\begin{gathered}
\bar{\mathbf u}_k^r
=
\left[
\frac{u_{k,1}^r}{u_{1,\max}^r},
\frac{u_{k,2}^r}{u_{2,\max}^r}
\right]^{\mathsf T},
\\
\mathcal L_{\mathrm{exp}}^r
=
\frac{1}{H}
\sum_{k=0}^{H-1}
\left\|
\bar{\mathbf u}_k^r
-
\bar{\mathbf u}_k^{r,\mathrm{exp}}
\right\|_2^2,
\end{gathered}
\end{equation}
where \(\mathbf u_k^r=[u_{k,1}^r,u_{k,2}^r]^{\mathsf T}\) denotes the platform-specific control. For differential-drive and Ackermann platforms, it corresponds to \([v,\omega]^{\mathsf T}\) and \([v,\delta]^{\mathsf T}\), respectively.

At training epoch $e$, the overall objective is
\begin{equation}
\begin{gathered}
\mathcal L_{\mathrm{train}}^{r,(e)}
=
\lambda_{\mathrm{exp}}
w(e)
\mathcal L_{\mathrm{exp}}^r
+
\mathcal L_{\mathrm{cost}}^r,
\\
w(e)
=
\frac{1}{
1+\exp\!\left[\beta(e-E_{\mathrm{half}})\right]
},
\label{eq:curriculum_loss}
\end{gathered}
\end{equation}
where $E_{\mathrm{half}}$ and $\beta>0$ control the midpoint and rate of expert-weight decay, respectively. At the beginning of training, $w(e)\approx1$, which suppresses zero-control collapse. As training proceeds, the expert constraint gradually weakens, allowing the policy to improve beyond expert imitation through task-level cost optimization. The curriculum expert-to-cost transfer is illustrated in Fig.~\ref{fig:method2}.

\subsection{Predictive Safety Lock for Hybrid Deployment}

To reduce deployment risks in abnormal environments, LECO performs a short-horizon safety check on the predicted neural trajectory before issuing control commands. A safety lock switches between the neural policy and the conventional hybrid A* and MPC branch.

At each control cycle, the first $M\leq H$ predicted control steps are rolled out using the kinematic model in Sec.~B. This defines a safety horizon of $T_{\mathrm{safe}}=M\Delta t$. We use $M=10$, resulting in $T_{\mathrm{safe}}=1.0\,\mathrm{s}$. Let $\rho_r$ denote the conservative equivalent radius of platform $r$. The minimum ESDF safety margin is defined as

\begin{equation}
\begin{gathered}
\Delta S_t^r
=
\min_{1\leq k\leq M}
\left[
\mathcal D_{\mathrm{ESDF}}
\left(
\hat{\mathbf p}_{t+k|t}^r
\right)
-
\rho_r
\right].
\end{gathered}
\end{equation}

The predicted trajectory is considered unsafe when $\Delta S_t^r<0$. Considering target validity and the safety-lock state, the neural branch is enabled only when
\begin{equation}
\begin{gathered}
\mathcal G_t^{\mathrm{neural}}
=
\mathbb I(V_t=1)\,
\mathbb I(t\geq t_{\mathrm{lock}})\,
\mathbb I(\Delta S_t^r\geq0),
\end{gathered}
\end{equation}
where \(V_t\in\{0,1\}\) denotes the validity state of the target prediction used at the current control cycle. The variable \(t_{\mathrm{lock}}\) denotes the earliest time at which the neural branch is allowed to resume. Let \(\boldsymbol\Gamma_t^r\) denote the reference path generated by hybrid A* for the classical branch. The final control command is given by

\begin{equation}
\mathbf u_t^{r,\mathrm{cmd}}
=
\begin{cases}
\hat{\mathbf u}_{t|t}^{r,\mathrm{neural}},
&
\mathcal G_t^{\mathrm{neural}}=1,
\\
\operatorname{MPC}
\left(
\mathbf x_t^r,
\boldsymbol\Gamma_t^r
\right),
&
\mathcal G_t^{\mathrm{neural}}=0.
\end{cases}
\end{equation}

When the predicted neural trajectory is unsafe, hybrid A* updates \(\boldsymbol\Gamma_t^r\) as an interim safe tracking path. If the target prediction becomes invalid, \(\boldsymbol\Gamma_t^r\) is generated as a reacquisition path using the latest valid target state. The resulting path also accounts for the desired tracking distance, line of sight, and obstacle clearance.

Once a risk event triggers the classical branch, the lock is updated as

\begin{equation}
t_{\mathrm{lock}}
=
t+N_{\mathrm{lock}}.
\end{equation}
where \(N_{\mathrm{lock}}\) denotes the minimum lock duration in control cycles. We set \(N_{\mathrm{lock}}=6\), corresponding to \(0.6\,\mathrm{s}\) at \(10\,\mathrm{Hz}\).

This keeps the classical branch active for at least $N_{\mathrm{lock}}$ control cycles and prevents frequent switching between the two branches. Neural control is resumed only when the target prediction becomes valid again, the safety lock has expired, and the short-horizon safety check is satisfied.

\begin{table}[t]
    \centering
    \vspace{2 mm}
    \caption{Comparison of Dynamic Target Tracking}
    \label{tab:tracking}
    \footnotesize
    \setlength{\tabcolsep}{2.2pt}
    \renewcommand{\arraystretch}{1.05}
    \begin{tabular}{c l c c c c}
        \toprule
        Scenario
        & Method
        & $E_v$ (m/s$^3$)
        & Occ. (\%)
        & Time (ms)
        & Succ. (\%) \\
        \midrule

        \multirow{5}{*}{$N_o=50$}
        & Elastic
        & 19.01
        & 3.02
        & \textbf{3.52}
        & 90 \\
        & YOPO
        & 43.21
        & 10.49
        & 17.73
        & 84 \\
        & A$^\ast$+MPC
        & 28.07
        & 0.24
        & 3.61
        & 96 \\
        & LECO (Abl.)
        & 18.44
        & 1.38
        & 3.90
        & 98 \\
        & LECO
        & \textbf{12.69}
        & \textbf{0.11}
        & 4.43
        & \textbf{100} \\
        \midrule

        \multirow{5}{*}{$N_o=100$}
        & Elastic
        & 33.29
        & 6.37
        & 4.61
        & 82 \\
        & YOPO
        & 67.25
        & 22.36
        & 19.93
        & 70 \\
        & A$^\ast$+MPC
        & 42.49
        & \textbf{2.84}
        & 5.12
        & 88 \\
        & LECO (Abl.)
        & 23.23
        & 5.58
        & 4.45
        & 84 \\
        & LECO
        & \textbf{19.66}
        & 3.33
        & \textbf{4.25}
        & \textbf{92} \\
        \midrule

        \multirow{5}{*}{$N_o=150$}
        & Elastic
        & 52.70
        & 8.76
        & \textbf{5.07}
        & 74 \\
        & YOPO
        & 80.32
        & 30.07
        & 23.61
        & 62 \\
        & A$^\ast$+MPC
        & 56.87
        & 13.80
        & 8.68
        & 80 \\
        & LECO (Abl.)
        & 32.18
        & 19.72
        & 6.90
        & 80 \\
        & LECO
        & \textbf{27.63}
        & \textbf{6.32}
        & 5.63
        & \textbf{86} \\
        \bottomrule
    \end{tabular}
    \vspace{-2 mm}
\end{table}

\section{Simulations and Benchmarking}

\subsection{Experimental Setup}

We evaluate LECO on a differential-drive robot and compare it with Elastic-Tracker~\cite{6}, YOPO-Planner~\cite{10}, and A$^\ast$+MPC. Since YOPO is designed for motion planning, we use the same tracking reference point as its planning goal without adding explicit target-visibility objectives. A$^\ast$+MPC also serves as the early-stage expert for LECO. LECO (Abl.) removes the predictive safety lock and stops the robot when obstacle clearance falls below the safety threshold.

Separate LECO policies are trained for the simulation, indoor, and outdoor settings using platform-specific dynamics and environment-specific simulation data. Each dataset contains 0.86M frames at 10\,Hz with a 9:1 training-validation split. Training uses 50 epochs and a batch size of 128 on one NVIDIA GeForce RTX 4090 D, taking 8.5\,h on average. Training and evaluation use disjoint maps and trajectories.

For simulation, we set $d_{\mathrm{vis}}=0.31\,\mathrm{m}$,
$d_{\mathrm{des}}=2.50\,\mathrm{m}$, $d_{\min}=0.70\,\mathrm{m}$,
$d_{\mathrm{los}}=0\,\mathrm{m}$, and
$d_{\mathrm{danger}}=0.31\,\mathrm{m}$. The map is
$80\,\mathrm{m}\times80\,\mathrm{m}$ with 50, 100, or 150 randomly generated static obstacles. Their length and width are independently sampled from $[0.5,18]\,\mathrm{m}$.

For each obstacle density, we conduct 50 independent continuous tracking trials. All methods use the same maps, target trajectories, initial states, and control frequency. We report linear jerk RMS $E_v$, occlusion rate, per-cycle computation time, and tracking success rate. $E_v$ denotes the RMS linear jerk computed from the executed velocity sequence. The occlusion rate measures the fraction of tracking time for which the target is not observable. A trial is successful if no target-loss event lasts longer than 4\,s.

\begin{figure}[t]
    \centering
    \vspace{2 mm}
    \includegraphics[width=\linewidth]{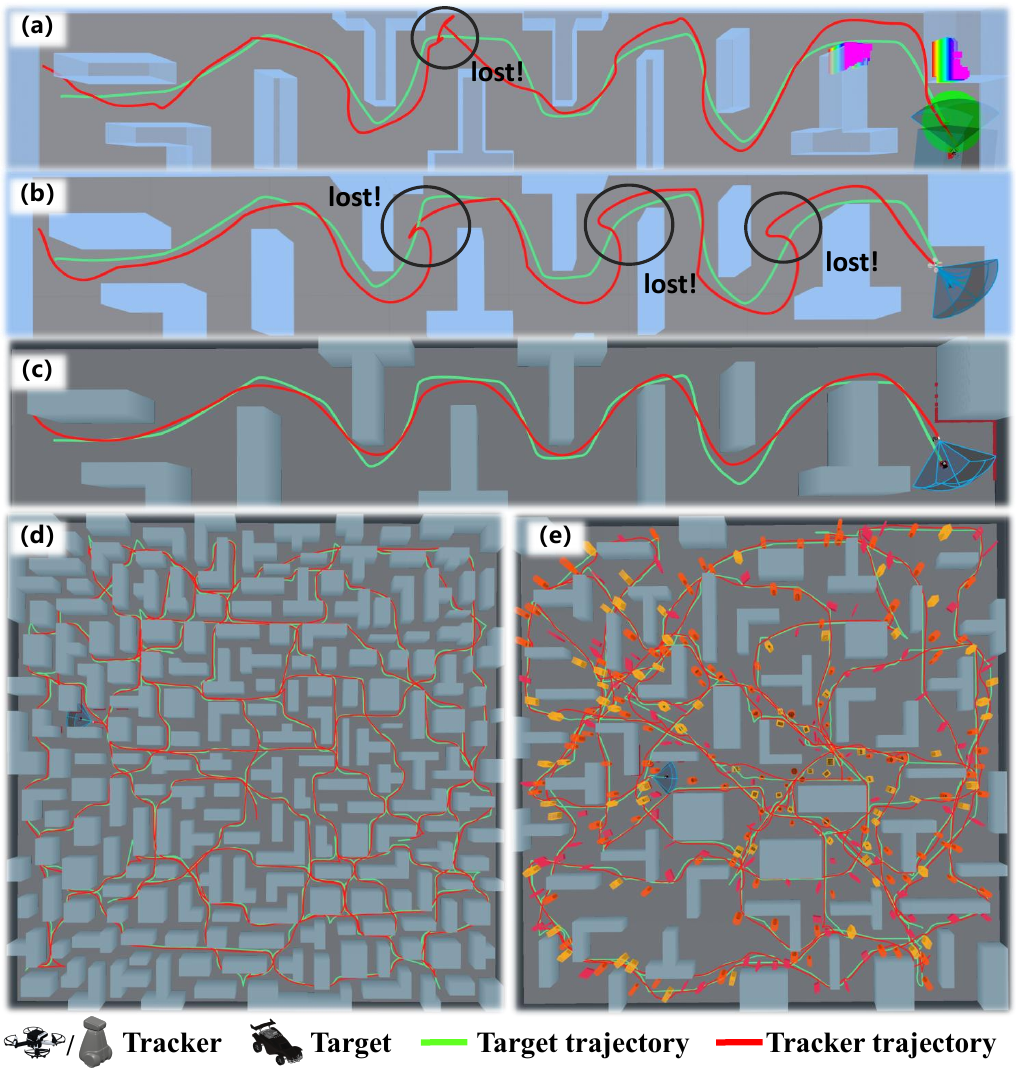}
    \caption{Qualitative tracking comparison and challenging scenarios. (a) Elastic-Tracker experiences one target-loss event. (b) YOPO experiences three target-loss events. (c) LECO maintains smooth tracking without target loss. (d) LECO tracking trajectories in Maze-200 with 200 dense obstacles, narrow passages, and consecutive turns. (e) LECO tracking trajectories in Maze-50 with 50 dense obstacles and temporary obstacle disturbances.}
    \vspace{-2 mm}
    \label{fig:exp1}
\end{figure}

\subsection{Comparative Evaluation}

As shown in Table~\ref{tab:tracking}, LECO achieves the lowest $E_v$ across all obstacle densities, reducing it by 33.2\%, 40.9\%, and 47.6\% over the best external baseline. It maintains 92\% and 86\% success with 100 and 150 obstacles, with only 6.32\% occlusion in the most challenging scene. Compared with LECO (Abl.), the full model consistently improves overall tracking success and reduces occlusion. At $N_o=150$, the occlusion rate drops from 19.72\% to 6.32\%, showing that predictive fallback better preserves tracking continuity than emergency stopping.

Fig.~\ref{fig:exp1}(a)--(c) shows one target-loss event for Elastic-Tracker and three for YOPO, while LECO maintains continuous tracking without target loss. YOPO performs worse in visibility because it does not explicitly optimize target observability. LECO also surpasses the A$^\ast$+MPC expert as decaying expert guidance enables further task-level optimization. Its 4.25--5.63\,ms runtime remains comparable to the optimization-based baselines; although Elastic-Tracker is slightly faster in two settings, LECO achieves substantially lower jerk and higher tracking success.

\subsection{Challenging Scenarios}

\begin{figure}[t]
    \centering
    \vspace{2 mm}
    \includegraphics[width=\linewidth]{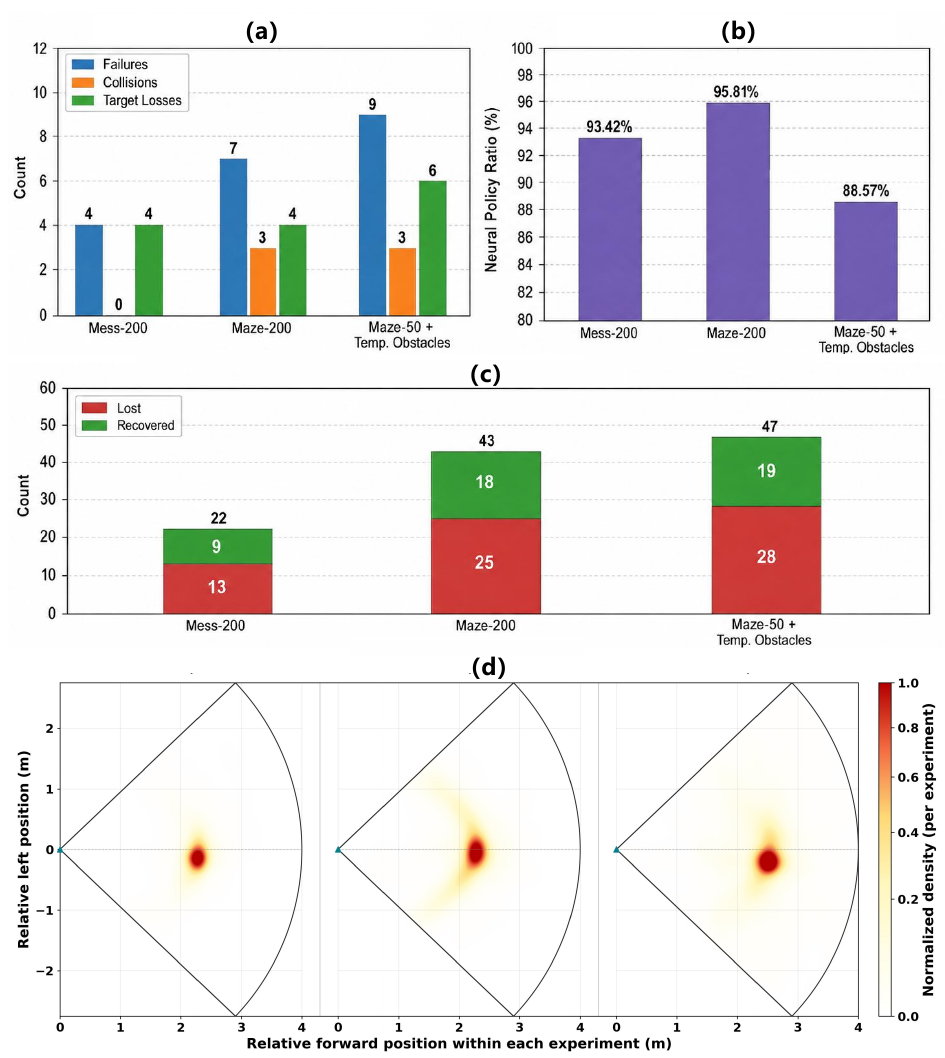}
    \caption{Quantitative evaluation in challenging scenarios. (a) Failures, collisions, and target-loss failures over 50 trials. (b) Neural-policy control ratio. (c) Target-loss and recovery events. (d) FoV heatmaps for Mess-200, Maze-200, and Maze-50 + Temporary Obstacles, from left to right.}
    \vspace{-2 mm}
    \label{fig:exp2}
\end{figure}

To further evaluate LECO under extreme conditions, we construct Mess-200, Maze-200, and Maze-50 + Temporary Obstacles. Each scenario contains 50 long-horizon trials with dense obstacles, narrow passages, sharp turns, or sudden disturbances. Representative trials are shown in Fig.~\ref{fig:exp1}(d)--(e), and the quantitative results are summarized in Fig.~\ref{fig:exp2}.

\begin{figure}[t] 

    \centering
    \includegraphics[width=\linewidth]{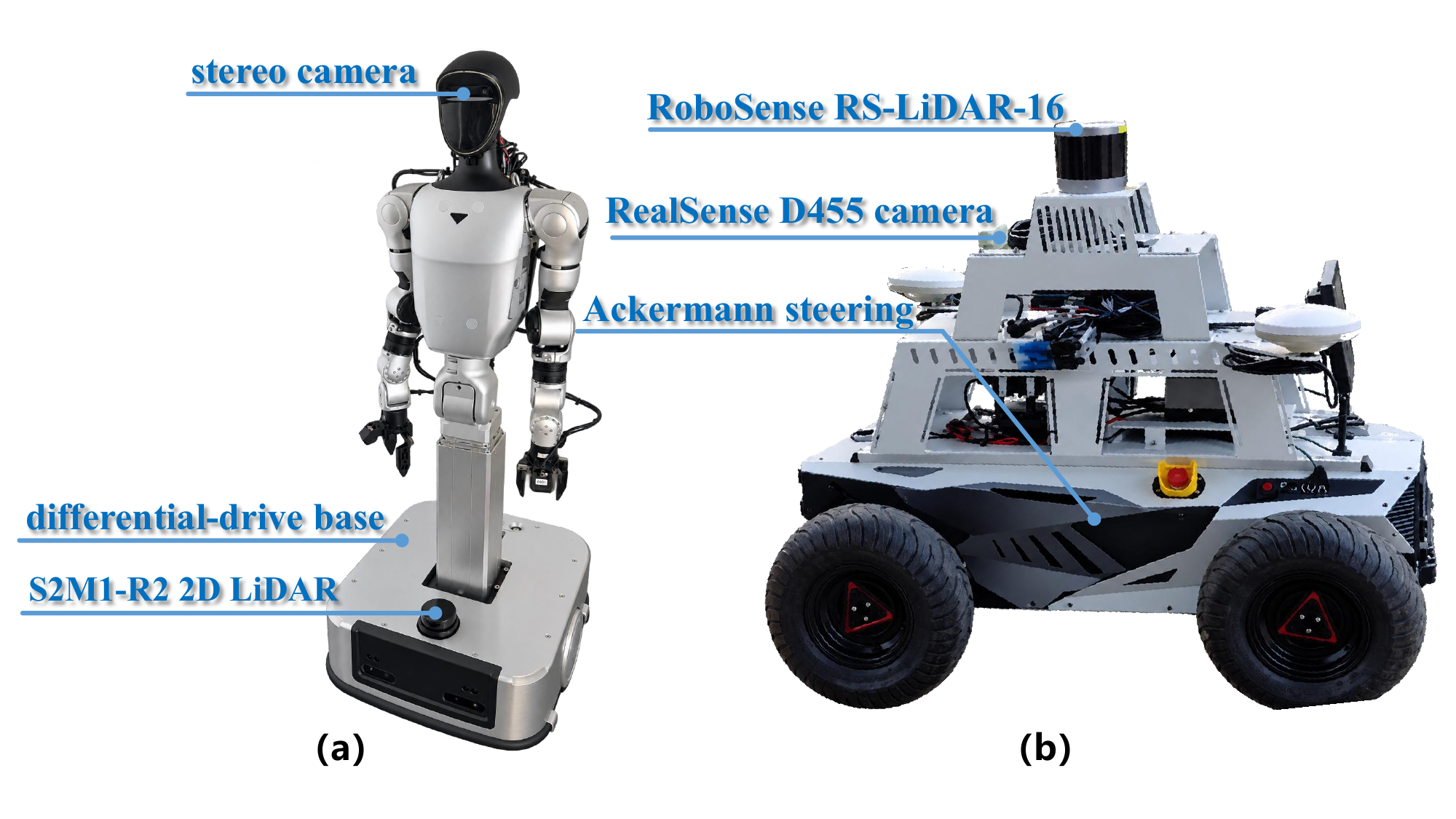}
    \caption{Robotic platforms used in the real-world experiments. (a) Unitree G1D with a differential-drive base, S2M1-R2 2D LiDAR, and stereo camera for indoor experiments. (b) AgileX HUNTER 2.0 with Ackermann steering, Intel RealSense D455 depth camera, and RoboSense RS-LiDAR-16 for outdoor experiments.}
    \vspace{-2 mm}
    \label{fig:robot}
\end{figure}

\begin{figure}[!ht]
    \centering
    \vspace{2 mm}
    \includegraphics[width=\linewidth]{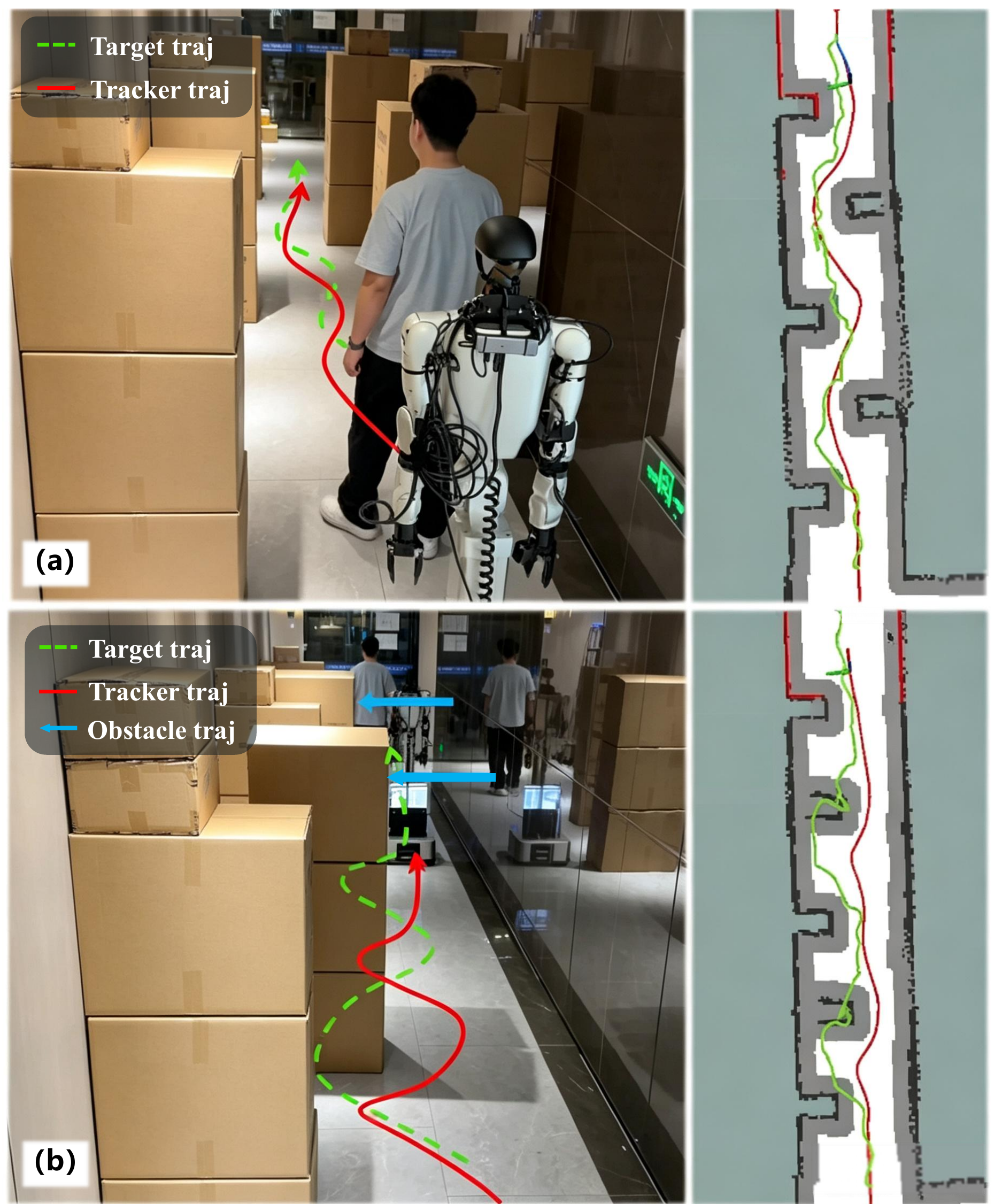}
    \caption{Real-world indoor target-tracking experiments. (a) Continuous tracking through an S-shaped narrow passage with consecutive turns and constrained free space. (b) Tracking under a temporary obstacle disturbance, demonstrating the response of the predictive safety lock.}
    \vspace{-2 mm}
    \label{fig:realworld1}
\end{figure}

Across the three scenarios, the neural policy accounts for 88.57--95.81\% of control decisions, showing that it remains the primary controller under challenging conditions. Mess-200 records no collision and only four failed trials, while Maze-200 still maintains a 95.81\% neural-policy ratio despite narrow passages and consecutive turns. Under temporary obstacle disturbances, the neural-policy ratio decreases to 88.57\%, indicating more frequent safety-lock intervention. LECO nevertheless maintains an 82\% success rate and recovers the target in 19 of 28 loss events. The FoV heatmaps further show that the target remains within the effective field of view for most of the tracking process.

\begin{figure}[t]
    \centering
    \vspace{2 mm}
    \includegraphics[width=\linewidth]{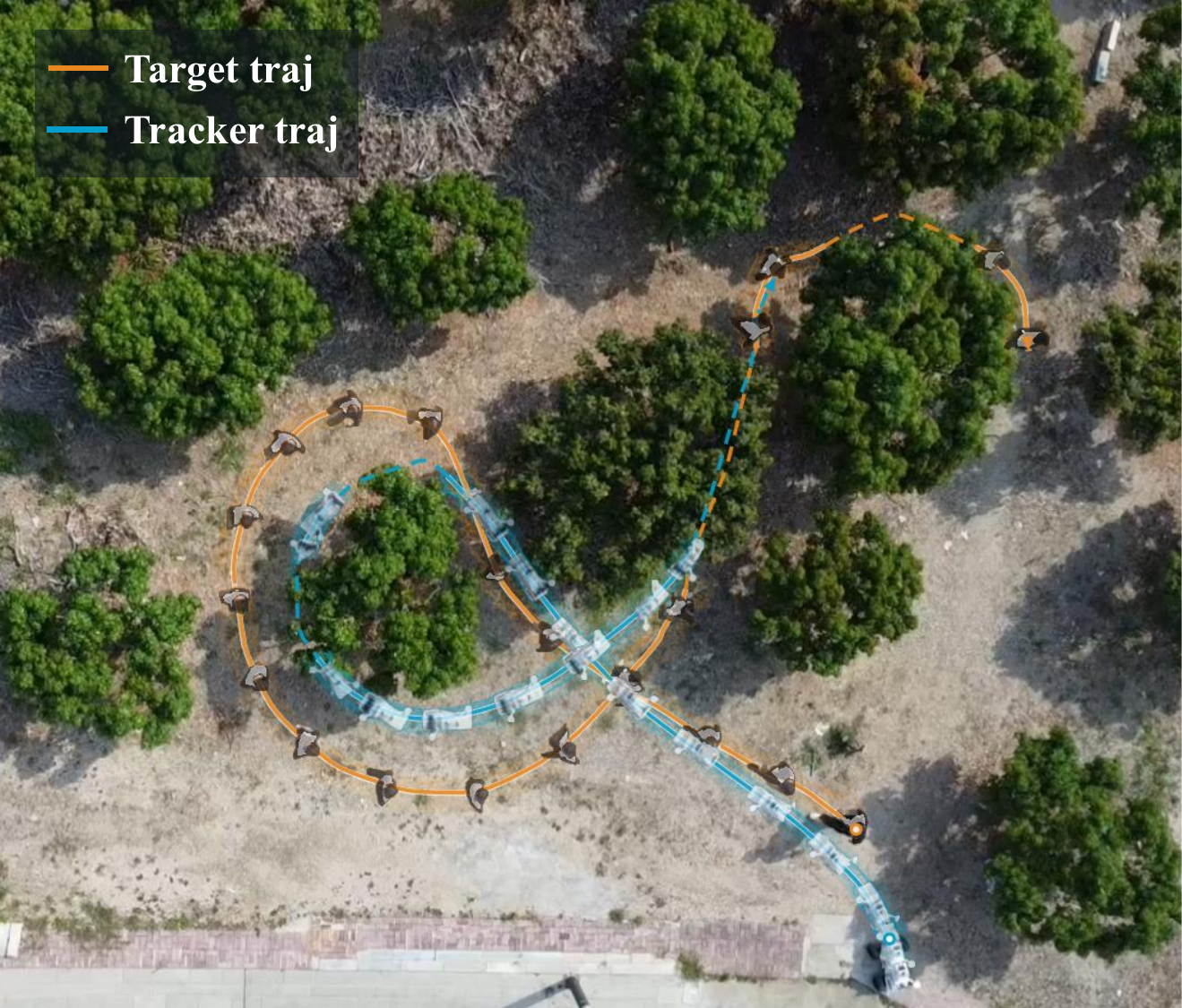}
    \caption{Long-range target tracking in a complex outdoor forest. The target (orange) and tracker (blue) trajectories demonstrate continuous tracking and obstacle avoidance through irregularly distributed natural obstacles.}
    \vspace{-2 mm}
    \label{fig:realworld2}
\end{figure}

\section{Real-World Experiments}

Real-world experiments are conducted in indoor cluttered environments and an outdoor forest environment using a differential-drive Unitree G1D and an Ackermann-steered AgileX HUNTER 2.0, respectively, as shown in Fig.~\ref{fig:robot}. We design three challenging scenarios, including temporary obstacle disturbance, continuous tracking in an S-shaped narrow passage, and long-range tracking in a complex forest. Separate policies are trained for the two platforms using their corresponding dynamics and control heads. These experiments evaluate tracking stability, obstacle avoidance, target observability, and long-horizon operation under different platform dynamics and environmental conditions.

\textbf{Continuous tracking in an S-shaped narrow passage.}
The target moves through a narrow multi-turn passage along an S-shaped or figure-eight route. This continuously changes the relative distance and bearing while introducing sharp turns, constrained free space, and short-term occlusions. The experiment mainly evaluates control smoothness, jerk, and target observability under complex geometric constraints. As shown in Fig.~\ref{fig:realworld1}(a), the robot completes long-range continuous tracking through the narrow passage while maintaining stable obstacle avoidance. The target also remains within the effective field of view for most of the tracking process, demonstrating stable tracking under consecutive turns and intermittent occlusions.

\textbf{Temporary obstacle disturbance.}
This experiment evaluates the response of the predictive safety lock to sudden environmental changes. As shown in Fig.~\ref{fig:realworld1}(b), a temporary obstacle is introduced during tracking and triggers the classical fallback when the predicted neural motion becomes unsafe. The fallback controller guides the robot around the obstacle while maintaining the tracking task. After the obstacle is bypassed and the safety condition is restored, control returns to the neural policy, allowing continuous tracking without prolonged interruption.

\textbf{Long-range tracking in a complex forest.}
The outdoor experiment evaluates long-term operation in an unstructured environment. It focuses on continuous target tracking, natural obstacle avoidance, and target reacquisition after short-term loss. As shown in Fig.~\ref{fig:realworld2}, the robot maintains stable tracking in a forest with irregular obstacle layouts and continuously changing free space. It remains responsive to local geometric changes while balancing obstacle avoidance and target retention. After temporary occlusion or target loss, the robot can recover the target and resume tracking, demonstrating stable long-horizon operation on the Ackermann platform.

\section{CONCLUSION AND FUTURE WORK}

This paper presented LECO, a unified framework for safe and efficient dynamic target tracking. It combines optimization-to-policy learning, kinematic-integrated control, and predictive safety lock. Together, these components improve control smoothness, tracking robustness, and runtime safety. Simulation results show improved tracking success, target visibility, and control smoothness with competitive millisecond-level computation, while real-world experiments demonstrate stable operation in indoor and outdoor environments. These advantages are maintained across dense obstacles, narrow passages, sudden disturbances, and unstructured environments overall. 

The current framework mainly considers planar kinematics and 2-D ESDF constraints. Future work will extend LECO to richer platform dynamics, 3-D environments, and more complex dynamic interactions.


\bibliographystyle{IEEEtran}
\bibliography{reference}

\end{document}